\documentclass[a4paper,fleqn]{cas-dc}

\usepackage[numbers]{natbib}
\usepackage{pifont}
\usepackage{url}
\ExplSyntaxOn

\ExplSyntaxOff

\begin{document}
\let\WriteBookmarks\relax
\def\floatpagepagefraction{1}
\def\textpagefraction{.001}
\shorttitle{Binaural Audio-Visual Instance Segmentation}
\shortauthors{S. Wang et~al.}

\title [mode = title]{Binaural Audio-Visual Instance Segmentation}

\author[1]{Saijun Wang}
\fnmark[1]
\ead{wangsaijun@tongji.edu.cn}
\author[1]{Guanfeng Tang}
\fnmark[1]
\ead{guanfeng.T@outlook.com}
\author[1]{Hongbo Zhao}
\ead{hongbozhao@tongji.edu.cn}
\author[1]{Zhicheng Lei}
\ead{2453748@tongji.edu.cn}
\author[1]{Yutong Zhang}
\ead{2551396@tongji.edu.cn}
\author[2]{Wei Ye}
\ead{yew@tongji.edu.cn}
\author[2]{Rui Fan}
\cormark[1]
\ead{rui.fan@ieee.org}

\fntext[1]{These authors contributed equally to this work.}
\cortext[1]{Corresponding author}
\nonumnote{This work was supported in part by the National Key R\&D Program of China under Grant 2025YFE0202900, in part by the National Natural Science Foundation of China under Grant 62473288 and Grant 62233013, in part by the State Key Laboratory of Human-Machine Hybrid Augmented Intelligence, Xi'an Jiaotong University (No. HMHAI-202406), in part by the Fundamental Research Funds for the Central Universities, in part by NIO University Programme (NIO UP), and in part by the Xiaomi Young Talents Program.}

\affiliation[1]{organization={College of Electronic and Information Engineering, Tongji University, Shanghai 201804, China}}
\affiliation[2]{organization={College of Electronic and Information Engineering, Shanghai Institute of Intelligent Science and Technology, State Key Laboratory of Autonomous Intelligent Unmanned Systems, Tongji University, Shanghai 201804, China}}

\begin{abstract}
Audio-visual segmentation (AVS) aims to segment sounding objects at the pixel level by integrating auditory and visual cues. However, existing methods are predominantly developed under the monaural setting and primarily rely on cross-modal semantic correspondence, which limits their ability to distinguish visually similar instances of the same semantic class. In contrast, humans naturally exploit binaural hearing, where interaural differences and direction-dependent acoustic filtering introduced by the head and pinnae provide physically grounded spatial cues for accurate sound source localization. Motivated by this observation, we introduce binaural audio-visual instance segmentation (BiAVIS), a new task that leverages synchronized binaural audio and video frames to segment sounding instances. To advance research on this task, we establish two benchmarks by manually annotating an existing binaural audio-visual dataset and collecting a new real-world dataset, BiAVIS-Bench, in more challenging and diverse scenarios. The proposed BiAVIS model (BiAVISM) features an audio-only sound source localization network that directly projects binaural spatial cues onto the image plane to learn spatial and semantic priors for sounding instances. A query-level audio-visual fusion strategy is subsequently introduced to inject these informative priors into the instance segmentation decoder. Extensive experiments conducted on the two proposed benchmarks demonstrate the superior performance of the BiAVISM over previous monaural AVS methods, especially in resolving instance-level intra-class ambiguity. On the more challenging BiAVIS-Bench, BiAVISM achieves 26.65 mAP and 45.90 FSLA, corresponding to relative improvements of approximately 35.35\% over AVISM in mean average precision~(mAP) and 11.62\% over ACVIS in frame-level sound localization accuracy~(FSLA). The source code will be released at \url{https://github.com/Saijun-Wang/BiAVIS}.
\end{abstract}


\maketitle

\section{Introduction}\label{sec.intro}
Vision and audition are two fundamental sensory modalities through which humans perceive and understand the world~\cite{garner2022cortical}. By integrating visual and auditory cues, people can robustly localize and identify sounding objects in complex scenarios~\cite{mazo2024auditory}. Inspired by this human perception mechanism, researchers have extensively explored a wide range of audio-visual tasks in recent years, spanning audio-visual recognition and reasoning~\cite{pr_av_action, pr_uc_avsr, pr_avqa}, audio-visual correspondence~\cite{arandjelovic2017look, arandjelovic2018objects} and sound source localization (SSL)~\cite{BAVNet, L2BNet}. More recently, audio-visual research has extended from coarse localization to pixel-level mask prediction of sounding objects, giving rise to audio-visual semantic segmentation~(AVSS)~\cite{avsbench} and audio-visual instance segmentation~(AVIS)~\cite{AVIS}.

\begin{figure*}
    \centering
    \includegraphics[width=1.0\linewidth]{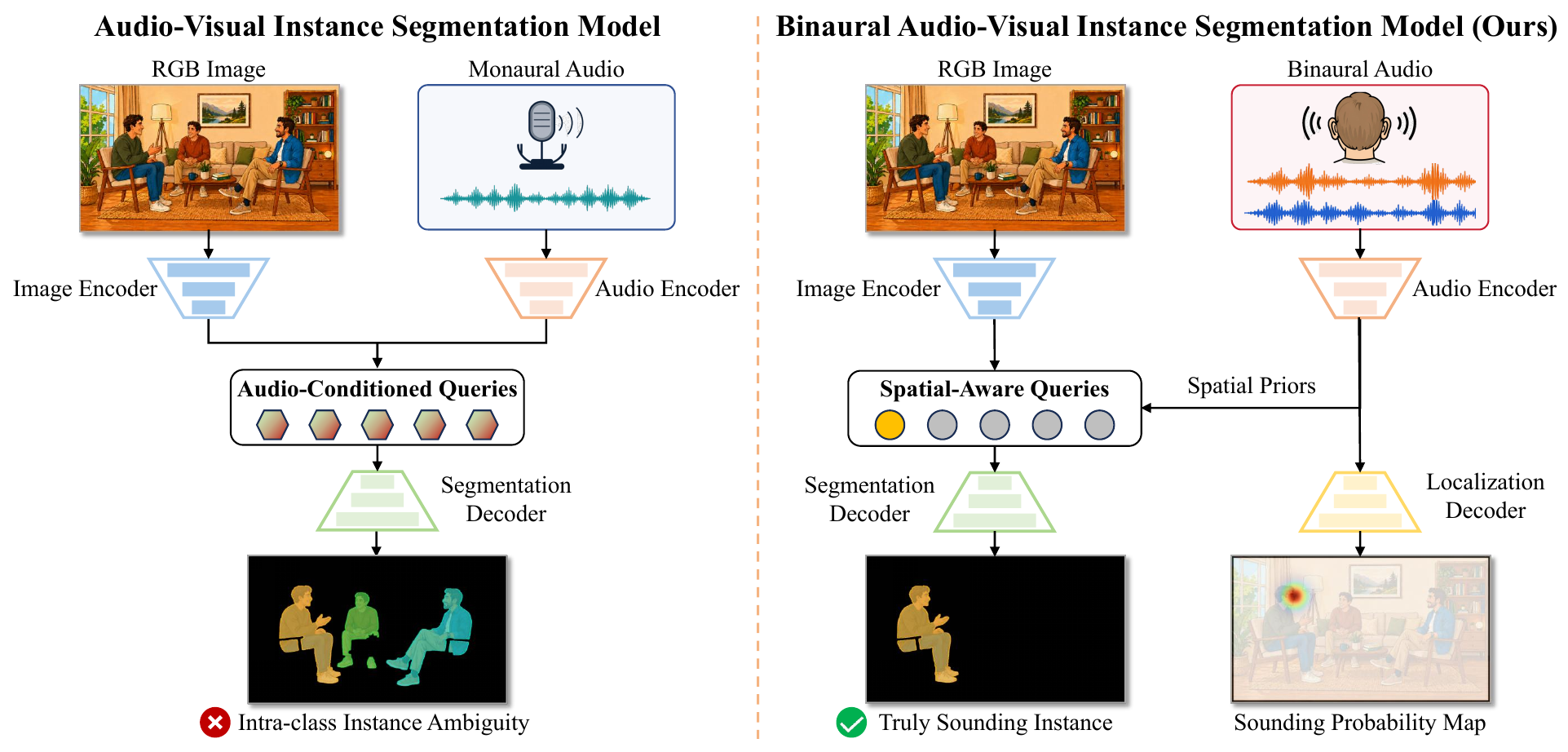}
    \caption{Comparison between AVISM and the proposed BiAVISM. AVISM conditions queries with monaural audio, leading to intra-class instance ambiguity. In contrast, BiAVISM exploits spatial priors, enabling segmentation of the truly sounding instance.}
    \label{fig_intro}
\end{figure*}

Nevertheless, a fundamental gap persists between human perception and current audio-visual segmentation (AVS) approaches. Humans naturally rely on binaural hearing to localize and distinguish sounding objects~\cite{van2019cortical}. Specifically, the auditory system exploits interaural time difference (ITD) and interaural level difference (ILD) between the two ears~\cite{middlebrooks1991sound}, while the head, torso, and pinnae further impose direction-dependent acoustic filtering on incoming sounds~\cite{wightman1989headphone}. Together, these cues provide physically grounded spatial information about the relative position of a sound source~\cite{francl2022deep}. In contrast, most existing AVS methods operate on monaural audio and primarily rely on cross-modal semantic correspondence between acoustic and visual representations of sounding objects~\cite{DDESeg}. For example, CAVP~\cite{chen2024unraveling} mines informative audio-visual contrastive pairs to better constrain cross-modal embedding learning, while AVSegFormer~\cite{AVSegFormer} introduces a transformer-based architecture with audio-conditioned queries to enhance cross-modal interaction. Despite their impressive performance on existing benchmarks, these monaural approaches often struggle in scenarios with multiple candidate instances that belong to the same semantic class, leading to severe intra-class ambiguity. As illustrated in Fig.~\ref{fig_intro}, when a group of people participates in a conversation, the active speaker can change over time. Existing monaural methods can associate audio with the general person category, but often fail to reliably distinguish the active speaker from silent participants. This limitation arises because monaural audio does not retain interaural differences, limiting access to spatial cues that can help distinguish visually similar instances.

To address this limitation, we introduce \textbf{Bi}naural \textbf{A}udio-\textbf{V}isual \textbf{I}nstance \textbf{S}egmentation (BiAVIS), a new task that leverages synchronized binaural audio and video frames to localize and segment sounding instances within the visual field. Unlike the monaural counterpart, binaural audio captured by human-like recording devices preserves physically grounded spatial cues associated with sound source direction, which can provide informative spatial priors for audio-visual segmentation. By exploiting these spatial cues, BiAVIS enables more reliable instance-level discrimination and helps resolve intra-class ambiguity in complex scenarios.
However, there is currently no publicly available dataset tailored to this task. This is mainly because conventional recording systems typically capture either monaural audio or two-channel audio without human-like head and pinna structures, neither of which can faithfully preserve the binaural spatial cues~\cite{song2026siren}. To mitigate this data scarcity, we first manually annotate FAIR-Play~\cite{FAIR-Play}, an existing binaural audio-visual dataset which was originally developed for binaural audio generation, with instance-level masks of sounding objects. Nevertheless, FAIR-Play is collected entirely in a single music room, resulting in extremely limited scene diversity. To further advance research on BiAVIS, we develop a customized data acquisition system comprising a binocular camera and a human-like binaural recording device. Based on this setup, we introduce BiAVIS-Bench, a challenging real-world benchmark covering diverse scenarios, sounding instances, and acoustic environments.

To fully exploit binaural audio, we develop an audio-only SSL network that learns both spatial and semantic priors to guide the subsequent instance segmentation network. Given only binaural audio as input, it is trained to simultaneously predict a dense sounding probability map and class probability maps on the image plane. A class-aware localization loss is further proposed to provide category-aware spatial supervision for the network. This formulation differs from existing binaural SSL methods, which typically rely on visual input or regress low-dimensional source directions, by predicting dense image-plane localization maps from binaural audio alone.

Building upon the proposed audio-only SSL network, a query-level audio-visual fusion strategy is introduced to inject binaural spatial and semantic priors into the instance segmentation decoder. Specifically, we leverage the intermediate features extracted from binaural audio by the SSL network to condition the object queries in the segmentation decoder. In this way, the object queries are conditioned on audio-derived spatial and semantic cues before interacting with visual features, guiding the decoder to generate audio-aware instance masks and class predictions. Additionally, an audio-visual consistency loss is introduced to align the predicted instance masks with the class-specific sounding probability maps produced by the SSL network. This loss encourages the segmentation decoder to focus on acoustically active regions and suppress visually salient but silent instances.
The main contributions are summarized as follows:
\begin{itemize}
    \item To the best of our knowledge, this is the first work to formulate binaural audio-visual instance segmentation (BiAVIS), which exploits human-like binaural spatial cues to resolve instance-level intra-class ambiguity that remains challenging for monaural AVIS methods.

    \item Two benchmarks are established for BiAVIS. FAIR-Play is annotated with instance-level sounding-object labels, and a new real-world benchmark, BiAVIS-Bench, is collected using a binocular camera and a human-like binaural recording device across diverse acoustic and visual environments.

    \item A dedicated audio-only binaural sound source localization network is designed to directly project binaural spatial cues onto the image plane as dense sounding and class probability maps. Unlike existing binaural localization methods that rely on visual input or regress low-dimensional source directions, the proposed network learns spatially grounded and category-aware representations from binaural audio alone.

    \item A class-aware audio-visual consistency objective is formulated to explicitly align instance-mask predictions with the audio-derived priors. Unlike existing AVIS methods that mainly rely on semantic audio conditioning, this objective strengthens the spatial alignment between the predicted instance masks and the binaural priors.

\end{itemize}
\section{Related Work}
This section reviews the two research directions most closely related to BiAVIS.
Section~\ref{sec.ssl} summarizes existing monaural and binaural sound source localization methods and discusses their limitations in spatial representation.
Section~\ref{sec.avs} reviews audio-visual segmentation methods and highlights the difficulty of existing monaural AVIS approaches in resolving instance-level intra-class ambiguity.
\subsection{Sound Source Localization}
\label{sec.ssl}
Sound source localization has been a long-standing topic which has received extensive attention in the computer vision and audio communities. Existing methods can be broadly categorized into three groups according to their input modalities. One line of work takes video frames and monaural audio as input, and typically formulates SSL as a cross-modal alignment problem~\cite{JSA,BetterAlign,OA-SSL}. For example, the study~\cite{senocak2018learning} develops an unsupervised two-stream framework that localizes sounding regions through audio-guided visual attention. In indoor environments, multimodal fusion has also been explored to combine complementary visual and acoustic cues for more robust sound source localization~\cite{pr_indoor_ssl}.
Despite their effectiveness, similar to existing AVS approaches, these monaural SSL methods also suffer from intra-class ambiguity, due to the lack of informative spatial cues for identifying the truly sounding instances. Another line of work incorporates binaural audio into the localization networks to exploit its inherent spatial cues for more accurate localization~\cite{L2BNet,CLUP}. For instance, the study~\cite{BAVNet} selectively fuses the image and binaural audio features through a recurrent neural network to produce a pixel-wise sounding probability map. More recently, audio--visual localization has also been investigated under challenging acoustic conditions; for example, Li et al.~\cite{pr_uav_avloc} jointly address audio denoising and localization under strong unmanned aerial vehicle ego-noise.
Although these approaches introduce binaural spatial cues into localization, their training and inference processes still rely heavily on visual input. As a result, they fail to fully exploit the binaural information. 
The third line of work relies solely on binaural audio for sound source localization, without incorporating visual information. Representative methods typically adopt a transformer-based architecture to extract and fuse features from the input audio, ultimately regressing the sound azimuth with respect to the binaural rig~\cite{BAST-mamba}. Nevertheless, such outputs are limited to estimating a single direction in the horizontal plane rather than producing pixel-wise localization maps, making them unsuitable for scenarios with multiple sound sources.
Existing binaural sound source localization methods either rely on visual input for image-plane localization or estimate low-dimensional source directions from binaural audio, and are therefore not designed to learn dense image-plane spatial and semantic representations directly from binaural audio.
\subsection{Audio-Visual Segmentation}
\label{sec.avs}
Audio-visual segmentation aims to localize and segment sounding objects by integrating auditory and visual cues. The study~\cite{avsbench} introduces the audio-visual semantic segmentation task and establishes the AVSBench dataset by collecting video clips from YouTube. To leverage audio semantics to guide image segmentation, this study designs a temporal, pixel-wise audio-visual interaction module that explores cross-modal correlations using a standard encoder-decoder architecture. Subsequent studies extend AVS to open-vocabulary settings by bridging vision--language and audio--language foundation models~\cite{pr_clipclap_avs}. 
Recently, inspired by DETR~\cite{detr} and Mask2Former~\cite{mask2former}, researchers have increasingly adopted query-based architectures for sounding object segmentation. For example, AVSegFormer~\cite{AVSegFormer} incorporates learnable queries that interact with audio features to capture relevant visual semantics, while COMBO~\cite{COMBO} explores multi-order bilateral relations across modality, temporal, and pixel levels to enhance the representation of object queries. 
LightAVSeg~\cite{LightAVSeg} improves the efficiency of audio-visual interaction without additional inference overhead.
Despite these advances in cross-modal interaction and efficiency, these methods remain formulated at the class level and cannot distinguish different sounding instances of the same category. To address this limitation, AVISM~\cite{AVIS} further extends the task to instance-level sounding object segmentation and tracking across video frames. H2S~\cite{H2S} further disentangles mixed audio and aligns the resulting acoustic streams with visual instances, while still relying on monaural input.
CTVIS~\cite{Ying_2023_ICCV} aligns training and inference for online association of visual identities across frames, whereas BiAVIS discriminates sounding instances within a frame using binaural spatial cues.
More recently, ACVIS~\cite{ACVIS} proposes a sound-aware ordinal counting loss that explicitly supervises the number of sounding instances via ordinal regression to prevent visual-only convergence during training. Existing AVIS methods mainly rely on audio-visual semantic correspondence to identify sounding instances, which remains insufficient for resolving multiple visually similar instances of the same category.
BiAVIS extends this setting by incorporating binaural spatial cues into audio-visual instance segmentation.
\section{Methodology}
\label{sec.method}
\begin{figure*}
    \centering
    \includegraphics[width=1\linewidth]{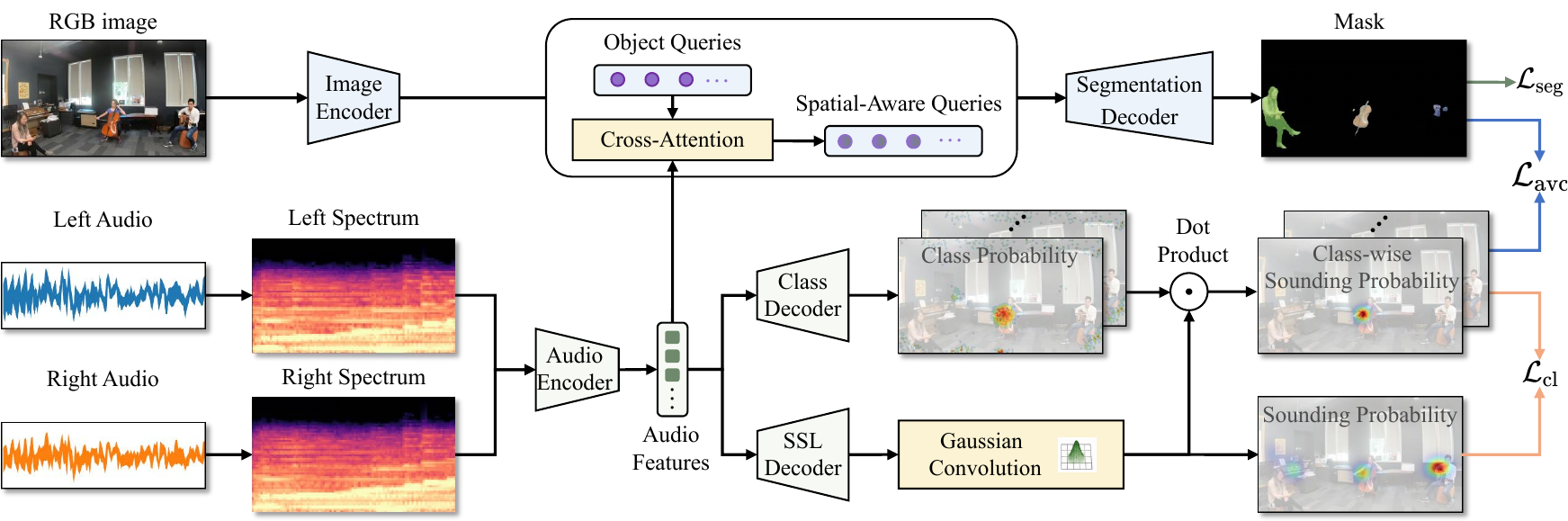}
    \caption{Overview of the proposed BiAVIS framework. The Audio Encoder denotes the complete binaural feature extraction pipeline, consisting of two channel-specific spectrogram encoders followed by a binaural-difference encoder. The audio-only SSL branch predicts sounding probability and class probability maps, while the visual branch follows a Mask2Former-style segmentation decoder. Binaural priors are injected via query-level cross-attention, where object queries are conditioned with binaural features to form audio-aware queries for segmentation prediction. The model is trained with segmentation, class-wise localization, and audio-visual consistency losses.}
    \label{fig.method}
\end{figure*}
This section presents the formulation and architecture of BiAVISM.
Section~\ref{sec.problem_definition} defines the BiAVIS task and its instance-level annotations.
Section~\ref{sec.binaural_ssl} describes the audio-only binaural sound source localization network for learning image-plane spatial and semantic priors.
Section~\ref{sec.av_fusion} presents the binaural-aware audio-visual fusion mechanism and the associated training objectives.
\subsection{Problem Definition}
\label{sec.problem_definition}
Fig.~\ref{fig.method} provides an overview of the proposed framework. We introduce binaural audio-visual instance segmentation~(BiAVIS), a new task that extends conventional AVIS from monaural audio to the binaural setting. 
Specifically, given an RGB image $\boldsymbol{I}\in\mathbb{R}^{H\times W\times 3}$ and its temporally synchronized left and right channel audio  $\boldsymbol{a}_\mathrm{l},\boldsymbol{a}_\mathrm{r}$, where $H$ and $W$ denote the image height and width and the three channels of $\boldsymbol{I}$ are RGB, BiAVIS aims to identify and segment all visible sounding instances. Let $N\geq 0$ denote the number of ground-truth sounding instances in the image, with the annotation set denoted by $\{(\boldsymbol{M}_i,c_i)\}_{i=1}^{N}$, where $\boldsymbol{M}_i\in\{0,1\}^{H\times W}$ is the binary mask of the $i$-th instance and $c_i\in\{1,\ldots,C\}$ is its category index among $C$ sounding-object categories. 
When $N=0$, the annotation set is empty, corresponding to a no-source sample. In addition, each ground-truth instance is associated with an image-plane sound source point $\boldsymbol{p}_i=(x_i,y_i)$ for localization supervision, where $x_i$ and $y_i$ are the horizontal and vertical pixel coordinates of the source point.
For human speakers, $\boldsymbol{p}_i$ approximates the mouth location and serves as a proxy for the speech source location~\cite{zhang2016mechanics}, while for other categories, it is set to the centroid of the instance mask.
BiAVIS outputs a set of predicted sounding instances $\{(\hat{\boldsymbol{M}}_i, \hat{c}_i)\}_{i=1}^{\hat{N}}$, where $\hat{N}\geq 0$ denotes the number of predicted sounding instances, $\hat{\boldsymbol{M}}_i$ denotes the $i$-th predicted mask and $\hat{c}_i$ denotes its corresponding class. When $\hat{N}=0$, the predicted set is empty, indicating that no sounding instance is detected.
\subsection{Binaural Sound Source Localization}
\label{sec.binaural_ssl}
To explicitly extract spatial priors from binaural audio, we design an audio-only sound source localization network that projects binaural spatial cues onto the image plane as sounding probability maps. This audio-to-image projection is enabled by the fixed relative pose between the camera and the binaural recording rig, which makes this projection consistent and learnable.

The binaural audio is first converted into log-mel spectrograms using the short-time Fourier transform, a mel-filterbank, and logarithmic compression.
Our audio encoder follows the MambaVision variant of BAST provided in the official BAST-Mamba implementation~\cite{BAST-mamba}, with the original localization prediction head removed.
Each spectrogram is partitioned into non-overlapping time--frequency patches, which are flattened and linearly projected into $D$-dimensional embeddings.
The embeddings are arranged in a fixed time-frequency order, preserving the local structure of the original binaural spectrograms. They are then fed into two architecturally identical but independently parameterized channel-specific spectrogram encoders, producing the left and right channel feature tensors $\boldsymbol{F}_{\mathrm{l}}$ and $\boldsymbol{F}_{\mathrm{r}}$.
Motivated by the fact that human sound localization relies on interaural differences, such as ITD and ILD, we feed these features into a binaural-difference encoder to extract spatial and semantic features.
The resulting fused binaural feature tensor is denoted by $\boldsymbol{F}_{\mathrm{b}}$ and passed to the subsequent segmentation network.

To supervise the localization network, we construct image-plane supervision from annotated source locations and class labels. A 2D Gaussian response with standard deviation $\sigma$ is centered at each annotated sound source location, and the responses are summed over the image to obtain the ground-truth map $\boldsymbol{G}$ at a resolution of $\frac{H}{8} \times \frac{W}{8}$. The map is subsequently normalized to obtain $\tilde{\boldsymbol{G}}$, which represents a smoothed sound source probability distribution.

On the prediction side, $\boldsymbol{F}_{\mathrm{b}}$ is pooled into a global embedding and passed to two lightweight feed-forward decoders. The SSL decoder predicts a non-negative sound source activation map $\boldsymbol{P} \in \mathbb{R}^{\frac{H}{8} \times \frac{W}{8}}$, where high responses indicate potential sound source locations on the image plane. Instead of directly predicting a dense Gaussian heatmap, the proposed method encourage the activation map to remain sparse, with most locations having near-zero responses and only a few source-related locations producing high responses. We then convolve the activation map with a Gaussian kernel using the same standard deviation as the target, producing the Gaussian-smoothed response map $\boldsymbol{H}$ and its normalized version $\tilde{\boldsymbol{H}}$. In this way, sparse activations are expanded into local Gaussian responses, yielding probability maps with Gaussian-smoothed response structures that are comparable to the target. 

To constrain both response magnitude and spatial distribution shape, we supervise the SSL decoder with a localization loss, which consists of a mean squared error~(MSE) loss $\mathcal{L}_{\mathrm{mse}}$ and a Jensen-Shannon~(JS) divergence~\cite{lin2002divergence} loss $\mathcal{L}_{\mathrm{js}}$. The MSE loss is computed between the unnormalized maps $\boldsymbol{H}$ and $\boldsymbol{G}$ to constrain pixel-wise response magnitude. In contrast, the JS loss aligns the spatial distribution shape of $\tilde{\boldsymbol{H}}$ with that of $\tilde{\boldsymbol{G}}$ while ignoring the overall response scale. 
This design restricts the effective prediction space by decoupling source activation prediction from Gaussian spatial smoothing. The decoder only needs to localize potential source centers, while the fixed Gaussian kernel determines the spatial extent of each response and provides a smoother objective than point-only supervision.

Although $\boldsymbol{G}$ provides spatial supervision for sound source locations, it is category-agnostic and cannot specify the category identity of each response peak. Existing AVIS methods commonly encode audio into a global embedding shared across the entire image, which may capture clip-level sound categories but does not associate them with specific spatial locations. This creates ambiguity in multi-source scenes: a location-only sounding probability map may highlight several source regions without distinguishing which category each peak belongs to.
To resolve this ambiguity, we design a class decoder for the SSL network, which predicts class logit maps $\boldsymbol{Y} \in \mathbb{R}^{C \times \frac{H}{8} \times \frac{W}{8}}$ over $C$ sounding-object categories. It is supervised by a class-wise Gaussian target heatmap $\boldsymbol{G}^{\mathrm{c}} \in \mathbb{R}^{C \times \frac{H}{8} \times \frac{W}{8}}$, where each annotated sound source point is used as the center of a Gaussian response in its corresponding class channel, following the same process as for $\boldsymbol{G}$. 
The class logit maps are converted into class probability maps $\boldsymbol{P}^{\mathrm{c}}$ by applying a softmax over the class dimension at each spatial location. We supervise the class probability maps with a Gaussian-weighted cross-entropy loss, which can be formulated as follows:
\begin{equation}
\mathcal{L}_{\mathrm{c}} = - \frac{\sum_{c,u,v}\boldsymbol{G}^{\mathrm{c}}(c,u,v) \log \boldsymbol{P}^\mathrm{c}(c,u,v)}{\sum_{c,u,v}\boldsymbol{G}^{\mathrm{c}}(c,u,v)}.
\end{equation}
Here, $c\in\{1,\ldots,C\}$ denotes the category index, and $(u,v)$ denotes a spatial location on the heatmap grid, with $u\in\{1,\ldots,H/8\}$ and $v\in\{1,\ldots,W/8\}$. The logarithm is the natural logarithm.
For samples without valid sound source annotations, $\boldsymbol{G}$ and $\boldsymbol{G}^{\mathrm{c}}$ are set to zero, since they are intensity targets. Accordingly, $\mathcal{L}_{\mathrm{c}}$ is set to zero. For the normalized target, $\tilde{\boldsymbol{G}}$ is a probability target and is set to a uniform distribution over the heatmap grid, so that it encodes no spatial preference.

The overall class-aware localization loss is defined as:
\begin{equation}
\mathcal{L}_{\mathrm{cl}}=\lambda_{\mathrm{js}}\mathcal{L}_{\mathrm{js}}+\lambda_{\mathrm{mse}}\mathcal{L}_{\mathrm{mse}}+\lambda_{\mathrm{c}}\mathcal{L}_{\mathrm{c}},
\end{equation}
where \(\lambda_{\mathrm{js}}\), \(\lambda_{\mathrm{mse}}\), and \(\lambda_{\mathrm{c}}\) are weighting coefficients that balance the contributions of the JS divergence loss, the MSE loss, and the classification loss, respectively.
\subsection{Binaural-Aware Audio-Visual Feature Fusion}
\label{sec.av_fusion}
For the visual branch, we follow the Mask2Former~\cite{mask2former} architecture. Given an input image, a visual backbone and a pixel decoder are used to extract multi-scale visual features. A transformer decoder then takes a fixed number of learnable object queries as input and decodes them into instance masks and class predictions, supervised by the segmentation loss \(\mathcal{L}_{\mathrm{seg}}\) after bipartite matching with ground-truth sounding instances.

To inject binaural priors into instance segmentation, we adopt a query-level audio--visual feature fusion mechanism. Instead of directly feeding the predicted heatmaps into the visual branch, we project the binaural features $\boldsymbol{F}_{\mathrm{b}}$ into the decoder feature space and use them as keys and values for cross-attention with the object queries, enabling each query to aggregate audio-derived spatial and semantic cues before visual decoding.
To further align the segmentation results with the audio priors, we introduce an audio-visual consistency loss.
Unlike cue transfer across label spaces~\cite{chen2022weakshot} or privileged modalities~\cite{chen2021depth}, these priors come from binaural audio at both training and inference.
The class-wise sounding probability maps $\boldsymbol{H}_c$ are defined as the element-wise product between the normalized sounding probability map $\tilde{\boldsymbol{H}}$ and the $c$-th channel of the class probability maps $\boldsymbol{P}^{\mathrm{c}}$.
This map represents the sounding probability that each location belongs to a sounding source of class $c$. Let \(\boldsymbol{M}_q\) denote the mask probability of the $q$-th query.
For any query $q$ and class $c$, we define the coverage score to measure how much of the class-wise sounding probability map is covered by the $q$-th predicted mask:
\begin{equation}
\mathrm{S}(\boldsymbol{M}_q,\boldsymbol{H}_c) = \frac{\sum_{u,v}\boldsymbol{M}_q(u,v)\boldsymbol{H}_c(u,v)}{\sum_{u,v}\boldsymbol{M}_q(u,v)}.
\end{equation}

Let $\mathcal{Q}^{+}$ denote the set of queries whose masks are matched to ground-truth instances and let $\mathcal{Q}^{-}$ denote the unmatched set. 
For a matched query \(q \in \mathcal{Q}^{+}\) assigned to a ground-truth instance of class $c_q$, we encourage its mask to cover the corresponding class-wise sounding probability map.
For an unmatched query \(q \in \mathcal{Q}^{-}\), we suppress its overlap with all class heatmaps by penalizing the maximum coverage score over classes, reducing dataset bias and improving instance selection in multi-source scenes.
Therefore, the audio-visual consistency loss is formulated as follows:
\begin{equation}
\begin{aligned}
\mathcal{L}_{\mathrm{avc}} ={}& \frac{1}{|\mathcal{Q}^{+}|+|\mathcal{Q}^{-}|}\Biggl(-\sum_{q \in \mathcal{Q}^{+}}\log\bigl(\mathrm{S}(\boldsymbol{M}_q,\boldsymbol{H}_{c_q})+\varepsilon\bigr)\\
&+\sum_{q \in \mathcal{Q}^{-}}\max_{c} \mathrm{S}(\boldsymbol{M}_q,\boldsymbol{H}_c)\Biggr).
\end{aligned}
\end{equation}
where $\varepsilon$ is a small positive constant added for numerical stability.

Finally, the complete framework is trained end-to-end using the following multi-task objective:
\begin{equation}
\mathcal{L} = \mathcal{L}_{\mathrm{seg}} + \lambda_{\mathrm{cl}} \mathcal{L}_{\mathrm{cl}} + \lambda_{\mathrm{avc}} \mathcal{L}_{\mathrm{avc}},
\end{equation}
where \(\lambda_{\mathrm{cl}}\) and \(\lambda_{\mathrm{avc}}\) balance the class-aware localization supervision and the audio-visual consistency loss, respectively. This design jointly optimizes instance-level segmentation, binaural SSL, and cross-modal consistency.
\begin{figure*}
    \centering
    \includegraphics[width=1\linewidth]{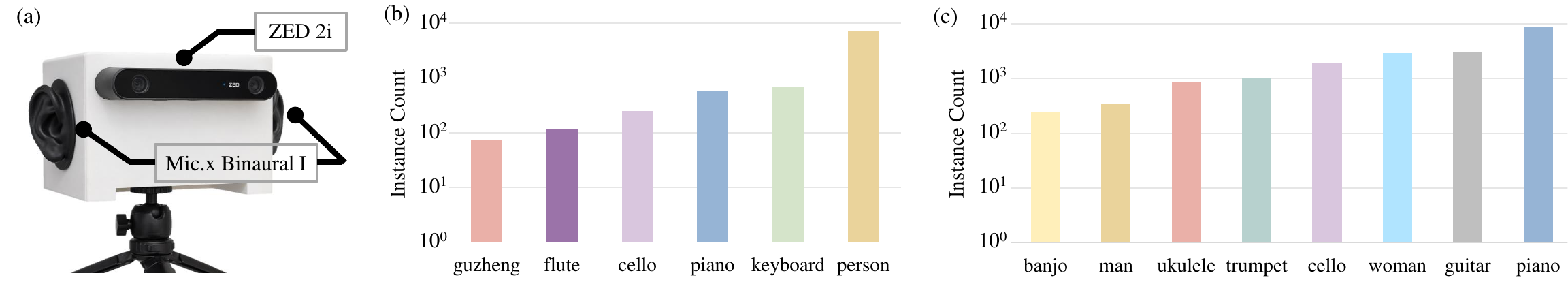}
    \caption{Dataset collection setup and category statistics.
(a) Our binaural audio-visual recording system, consisting of a ZED 2i binocular camera and a Mic.x Binaural I microphone mounted on a custom rig.
(b) Instance counts per category in BiAVIS-Bench.
(c) Instance counts per category in FAIR-Play.}
    \label{fig.dataset}
\end{figure*}

\section{Experiments}
\label{sec.experiments}
This section provides a comprehensive experimental evaluation of BiAVISM.
Section~\ref{sec.datasets} introduces the two benchmarks and their annotation protocols.
Section~\ref{sec.exp_setup} describes the evaluation metrics, baseline settings, and implementation details.
Section~\ref{sec.comparison} compares BiAVISM with representative methods from related tasks.
Section~\ref{sec.ablation} analyzes the contributions of the proposed components and binaural cues.
Section~\ref{sec.limitation} discusses the limitations of the current formulation.
\subsection{Datasets}
\label{sec.datasets}

We evaluate BiAVISM on two datasets: FAIR-Play~\cite{FAIR-Play} and a newly collected binaural audio-visual dataset, BiAVIS-Bench.
FAIR-Play is a public dataset that pairs videos with binaural audio recorded using a fixed custom rig. However, it does not provide the sounding-object annotations required for BiAVIS. To this end, we annotated FAIR-Play with pixel-wise sounding instance masks, semantic labels, and sound source points following the protocol described in the supplementary material. As shown in Fig.~\ref{fig.dataset}, the annotated FAIR-Play dataset comprises 1,364 videos and 13,304 clips, covering 8 object categories and 19,164 annotated sounding instances.

Nevertheless, FAIR-Play is collected entirely in a single music room, resulting in extremely limited scene diversity. To better evaluate BiAVISM in more challenging and diverse scenarios, we built a customized data acquisition system and collected a new large-scale benchmark. The system consists of a ZED 2i binocular camera and a Mic.x Binaural I microphone, rigidly mounted on a 3D-printed rig that approximates the relative geometry of human eyes and ears. The camera records videos at 30 fps with a resolution of $1{,}280 \times 720$ pixels, while the binaural microphone captures two-channel audio at 48 kHz and 32-bit depth. Audio and video streams are soft-synchronized via a shared ROS2 trigger and temporally aligned using timestamps, with a synchronization error bounded by $10.4$~$\mu s$. Compared with the FAIR-Play dataset, the proposed BiAVIS-Bench covers a more diverse range of scenes, including indoor, outdoor, and semi-open hall environments. Additionally, it places particular emphasis on scenarios involving multiple instances of the same category. The collected BiAVIS-Bench contains 870 videos and 8,052 clips, covering 6 object categories and 8,656 annotated sounding instances.

For both datasets, videos are divided into one-second clips, and only the center frame of each clip is used for the BiAVIS task. The temporal scope of these one-second clips is discussed in the supplementary material. Each sounding instance is annotated with a pixel-level mask, a semantic category label, and a sound source point. Each clip contains up to four simultaneously sounding instances, covering no-source, single-source, and multi-source cases. The training and test splits are constructed at the video level with a ratio of 7:3, ensuring balanced distributions of object categories and instance counts.

Further details on ethics and privacy considerations for BiAVIS-Bench are provided in the supplementary material.
\subsection{Experimental Setup}
\label{sec.exp_setup}
Following AVIS~\cite{AVIS}, we report mean average precision~(mAP) and frame-level sound localization accuracy~(FSLA) on both FAIR-Play and BiAVIS-Bench to evaluate the segmentation and localization performance of our method against representative approaches. 
Specifically, mAP evaluates the overall instance-level segmentation quality of sounding objects, while FSLA measures the proportion of frames in which sounding objects are correctly predicted in terms of category and mask overlap. FSLA further includes three sub-metrics: FSLAn, FSLAs, and FSLAm, which evaluate performance in the no-source, single-source, and multi-source cases, respectively. Higher values indicate better performance for all these metrics. 
The methods compared span three related tasks: video instance segmentation (VIS)~\cite{mask2former, MP-Former, PolarNeXt}, audio-visual semantic segmentation (AVSS)~\cite{COMBO, AVSegFormer, avsbench, DDESeg}, and audio-visual instance segmentation (AVIS)~\cite{AVIS, ACVIS, H2S}. Since VIS methods do not take audio as input, they serve as vision-only instance segmentation baselines. For AVSS methods based on query-based mask prediction, we use their query-level mask outputs to obtain instance-level predictions without changing the model architecture, loss functions, or training procedure.
For all methods originally designed for monaural audio, we convert each binaural recording into a single-channel waveform by averaging the synchronized left and right channels.
The resulting waveform is then processed using the original audio preprocessing pipeline and model architecture of each baseline.
For a fair comparison, all methods in Tables~\ref{tab.segmentation} use the same ImageNet-pretrained ResNet-50~\cite{ResNet-50} as the visual backbone, and are trained with the same batch size of 8 and a maximum training budget of 10 epochs. Both the AVSS and AVIS methods adopt Mask2Former-style query-based Transformer decoders, with the primary differences lying in how audio features are extracted, integrated with visual features, and used to guide mask prediction. Training beyond 10 epochs yields no further gains, indicating that all methods have converged under this setting.
Quantitative comparisons with representative sound source localization methods, using Pearson’s correlation coefficient (CC), Similarity Metric (SIM), and Earth Mover's Distance (EMD), are reported in the supplementary material.

For implementation, we adopt the MambaVision variant of BAST~\cite{BAST-mamba} as the audio encoder. All parameters of the audio encoder are jointly optimized with the sound source localization and instance segmentation branches during end-to-end training. The loss weights are set to $\lambda_{\mathrm{js}}=1.0$, $\lambda_{\mathrm{mse}}=1.0$, $\lambda_{\mathrm{c}}=0.3$,  $\lambda_{\mathrm{cl}}=1.2$, and $\lambda_{\mathrm{avc}}=0.5$. All models are optimized using AdamW~\cite{adamw} on a single NVIDIA GeForce RTX 4090 GPU. During training, input images are resized to $640 \times 360$ pixels, while evaluation is conducted at the original resolution. For efficiency, single-frame inference at $1{,}280 \times 720$ pixels has a median latency of about 145 ms and a peak PyTorch memory allocation of about 2.6 GiB over 30 timed runs. Training on FAIR-Play and BiAVIS-Bench takes approximately 3 and 2 hours, respectively. The complete preprocessing, architecture, and training parameters are listed in the supplementary material.

\begin{figure*}
    \centering
    \includegraphics[width=1\textwidth]{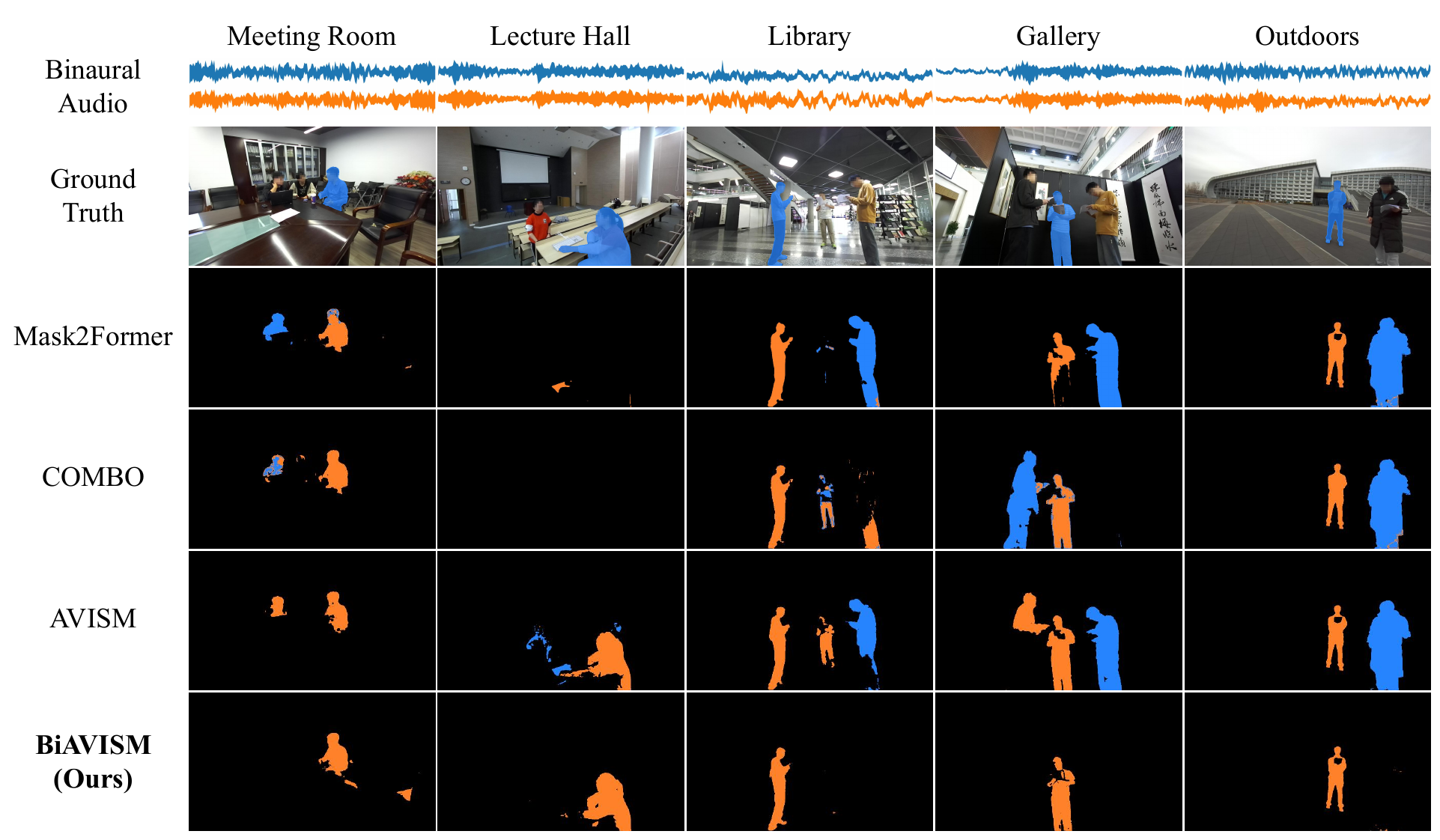}
    \caption{Qualitative comparison of sounding-object segmentation results across diverse scenes.}
    \label{fig.qualitative}
\end{figure*}
\subsection{Comparison with Methods from Related Tasks}
\label{sec.comparison}
\begin{table*}
  \caption{Quantitative comparison of different models from related tasks on the FAIR-Play and BiAVIS-Bench test sets.}
  \label{tab.segmentation}
  \centering
  \renewcommand{\arraystretch}{1.20}
  \setlength{\tabcolsep}{0.8mm}
  \begin{tabular}{clccccccccccc}
    \toprule
    \multirow{2}{*}{Task} & \multirow{2}{*}{Networks} & \multirow{2}{*}{Audio} & \multicolumn{5}{c}{FAIR-Play} & \multicolumn{5}{c}{BiAVIS-Bench} \\
    \cmidrule(lr){4-8} \cmidrule(lr){9-13}
    & & & mAP & FSLA & FSLAn & FSLAs & FSLAm & mAP & FSLA & FSLAn & FSLAs & FSLAm \\
    \midrule
    \multirow{3}{*}{VIS}
    & Mask2Former~\cite{mask2former} \textcolor{gray}{\tiny (CVPR'22)} &\ding{55} &35.71 &60.56 &3.52 &72.38 &44.23 &22.32 &40.22 &10.58 &45.07 &16.74 \\
    & MP-Former~\cite{MP-Former} \textcolor{gray}{\tiny (CVPR'23)} &\ding{55} &26.43 &54.17 &0.00 &\textbf{73.98} &24.40 &18.89 &19.82 &2.88 &21.95 &10.89 \\
    & PolarNeXt~\cite{PolarNeXt} \textcolor{gray}{\tiny (CVPR'25)} &\ding{55} &30.41 &36.55 &0.00 &51.12 &14.45 &15.72 &28.61 &2.88 &33.32 &4.71 \\
    \midrule
    \multirow{3}{*}{AVSS}
    & COMBO~\cite{COMBO} \textcolor{gray}{\tiny (CVPR'24)} &\checkmark &33.27 &56.27 &17.65 &63.86 &45.91 &19.00 &38.49 &9.62 &43.46 &13.88 \\
    & AVSegFormer~\cite{AVSegFormer} \textcolor{gray}{\tiny (AAAI'24)} &\checkmark &27.36 &44.88 &76.47 &59.75 &18.36 &15.47 &28.61 &4.81 &32.30 &11.15 \\
    & AVS-Semantic~\cite{avsbench} \textcolor{gray}{\tiny (IJCV'25)} &\checkmark &34.04 &53.66 &15.29 &65.11 &36.86 &20.14 &36.57 &7.69 &41.51 &12.20 \\
    & DDESeg~\cite{DDESeg} \textcolor{gray}{\tiny (CVPR'25)} & \checkmark & 8.04 & 35.92 & 25.88 & 42.98 & 24.79 & 9.89 & 21.10 & 3.85 & 28.85 & 5.37 \\
    \midrule
    \multirow{3}{*}{AVIS}
    & AVISM~\cite{AVIS} \textcolor{gray}{\tiny (CVPR'25)} &\checkmark &33.68 &56.94 &56.47 &66.17 &41.63 &19.69 &40.23 &5.77 &45.13 &18.08 \\
    & ACVIS~\cite{ACVIS} \textcolor{gray}{\tiny (ICASSP'26)} &\checkmark &32.75 &60.68 &4.71 &70.18 &48.14 &18.35 &41.12 &5.77 &46.04 &19.26 \\
    & H2S~\cite{H2S} \textcolor{gray}{\tiny (ACM MM'26)} & \checkmark & 11.99 & 49.84 & 17.65 & 64.26 & 27.74 & 12.46 & 40.48 & 9.62 & 45.09 & 19.16 \\
    & \textbf{BiAVISM (Ours)} &\checkmark &\textbf{37.63} &\textbf{62.90} &\textbf{89.41} &68.63 &\textbf{51.85} & \textbf{26.65} & \textbf{45.90} & \textbf{60.58} & \textbf{48.14} & \textbf{24.78} \\
    \bottomrule
  \end{tabular}
\end{table*}

The qualitative comparisons across diverse scenes are shown in Fig.~\ref{fig.qualitative}, while the quantitative segmentation results are reported in Tables~\ref{tab.segmentation}.
As shown in Fig.~\ref{fig.qualitative}, competing methods often suffer from intra-class ambiguity, mistakenly segmenting visually salient but non-sounding instances from the same semantic category. In contrast, BiAVISM produces masks that are more consistent with the annotated sounding instances across indoor and outdoor scenarios, suggesting that binaural spatial cues help suppress false activations on visually salient but silent instances.
As presented in Tables~\ref{tab.segmentation}, BiAVISM achieves the best overall mAP and FSLA on both datasets and consistently outperforms the compared monaural AVIS methods, with more pronounced gains on BiAVIS-Bench, which contains more challenging scenes with multiple visually similar instances of the same category. These results demonstrate that the proposed approach benefits from explicitly exploiting binaural spatial cues in addition to audio-visual semantic correspondence, enabling better discrimination of sounding instances than monaural AVIS methods that mainly rely on semantic information.
Although our model does not achieve the highest FSLAs on FAIR-Play, this is likely due to the dataset's strong single-source bias. When only one object is sounding, it is often also the most visually salient object, allowing visual-only or weakly audio-conditioned baselines to achieve high single-source scores. However, these methods perform much worse in no-source cases, suggesting they are more prone to false positives for visually salient but non-sounding objects.
Compared with FAIR-Play, BiAVIS-Bench intentionally includes more challenging scenes with multiple salient objects, especially same-category instances. The advantage of BiAVISM, therefore, becomes more evident on this benchmark, where the model outperforms the best-performing monaural baselines by 32.32\% in mAP and 11.62\% in FSLA. 
Overall, these results suggest that extending AVIS from monaural audio to binaural audio improves the model's spatial awareness of sounding instances, and that BiAVIS-Bench serves as a more challenging benchmark for evaluating this ability. A three-seed evaluation of the FAIR-Play comparison is reported in the supplementary material.

Despite using no image input, the proposed localization network outperforms multimodal methods that take images as additional input. On BiAVIS-Bench, it yields relative improvements of 19.35\% in CC and 50.00\% in SIM, together with a 42.60\% reduction in EMD. Full quantitative comparisons and qualitative localization results are provided in the supplementary material.

\subsection{Ablation Studies}
\label{sec.ablation}
This subsection analyzes the contributions of the key components of BiAVISM from two aspects.
Section~\ref{sec.ablation_loss} evaluates the effects of the Gaussian rendering, the JS divergence losses, the class-aware localization, and the audio-visual consistency.
Section~\ref{sec.ablation_binaural} further isolates the contribution of binaural spatial cues by comparing the original binaural input with different monauralized variants.
\subsubsection{Effect of Localization and Consistency Losses}
\label{sec.ablation_loss}
\begin{table}
  \caption{Ablation study on the proposed localization loss conducted on the BiAVIS-Bench dataset.}
  \label{tab.ablation_loc}
  \centering
  \renewcommand{\arraystretch}{1.15}
  \setlength{\tabcolsep}{2.2pt}
  \begin{tabular}{cccccc}
    \toprule
    Gaussian & $\mathcal{L}_{\mathrm{mse}}$ & $\mathcal{L}_{\mathrm{js}}$
    & CC $\uparrow$ & SIM $\uparrow$ & EMD $\downarrow$ \\
    \midrule
    \ding{55} & \checkmark & \ding{55} & 0.01 & 0.02 & 5.35 \\
    \ding{55} & \checkmark & \checkmark & 0.06 & 0.05 & 4.20 \\
    \checkmark & \checkmark & \ding{55} & 0.31 & 0.24 & 3.41 \\
    \checkmark & \checkmark & \checkmark & \textbf{0.37} & \textbf{0.33} & \textbf{2.87} \\
    \bottomrule
  \end{tabular}
\end{table}

\begin{table}
  \caption{Ablation study on the proposed class and audio-visual consistency loss conducted on the BiAVIS-Bench dataset.}
  \label{tab.ablation_obj}
  \centering
  \renewcommand{\arraystretch}{1.15}
  \setlength{\tabcolsep}{1.8pt}
  \begin{tabular}{ccccccc}
    \toprule
    $\mathcal{L}_{\mathrm{c}}$ & $\mathcal{L}_{\mathrm{avc}}$
    & mAP $\uparrow$ & FSLA $\uparrow$ & FSLAn $\uparrow$ & FSLAs $\uparrow$ & FSLAm $\uparrow$ \\
    \midrule
    \ding{55} & \ding{55} & 21.65 & 30.90 & 0.00 & 34.09 & 19.55 \\
    \ding{55} & \checkmark & 13.12 & 33.58 & 4.81 & 37.76 & 14.51 \\
    \checkmark & \ding{55} & 20.81 & 32.73 & 3.85 & 36.46 & 16.87 \\
    \checkmark & \checkmark & \textbf{26.65} & \textbf{45.90} & \textbf{60.58} & \textbf{48.14} & \textbf{24.78} \\
    \bottomrule
  \end{tabular}
\end{table}
We analyze the effect of localization supervision in Table~\ref{tab.ablation_loc} and the effect of class and audio-visual consistency losses in Table~\ref{tab.ablation_obj}. 
For localization, directly supervising the localization map without Gaussian rendering yields poor alignment with the target distribution, as evidenced by near-zero CC and SIM scores. This shows that directly predicting a dense heatmap is difficult, since the model needs to learn both the sound source activation locations and the spatial extent of each response. 
With Gaussian rendering, the model only needs to predict sparse source activations, while the fixed Gaussian kernel determines the local spatial extent.
The best performance is achieved when Gaussian rendering, $\mathcal{L}_{\mathrm{mse}}$, and $\mathcal{L}_{\mathrm{js}}$ are used together. This verifies that $\mathcal{L}_{\mathrm{mse}}$ and $\mathcal{L}_{\mathrm{js}}$ provide complementary supervision, with the former constraining the response magnitude and the latter aligning the normalized spatial distribution.

For class and audio-visual consistency losses, using either loss alone improves FSLA only moderately, with improvements of up to 8.67\% with $\mathcal{L}_{\mathrm{avc}}$ and 5.92\% with $\mathcal{L}_{\mathrm{c}}$. 
Moreover, $\mathcal{L}_{\mathrm{avc}}$ alone even reduces mAP, indicating that the localization map is category-agnostic and cannot reliably indicate which sound category each response belongs to. 
When $\mathcal{L}_{\mathrm{c}}$ and $\mathcal{L}_{\mathrm{avc}}$ are used together, the model achieves the best overall performance, improving mAP by 23.09\% and FSLA by 48.54\%. These results validate jointly learning spatial localization and audio category prediction.

\subsubsection{Contribution of Binaural Cues}
\label{sec.ablation_binaural}
To quantify the contribution of binaural cues, we compare the original binaural input with three monauralized variants in Table~\ref{tab.ablation_binaural}.
The avg setting averages the two channels and duplicates the averaged signal to both channels, while L/L and R/R duplicate the left and right channel respectively, across both input channels; L/R denotes the original binaural input.
All variants are trained with the same architecture and training protocol, with only the audio input construction changed.
Duplicating or averaging channels removes interaural differences while preserving monaural spectral and semantic information.
Compared with the strongest monauralized variant, the original binaural input improves FSLA and FSLAm by 7.69 and 12.83 points on FAIR-Play, and by 2.83 and 5.95 points on BiAVIS-Bench, respectively.
The improvement is most pronounced on FSLAm, which evaluates multi-source scenarios that require instance-level spatial discrimination.
These results indicate that the performance gain stems from exploiting interaural spatial cues, rather than merely from additional audio information or model capacity.

\begin{table*}
  \caption{Ablation study on binaural versus monauralized audio inputs.}
  \label{tab.ablation_binaural}
  \centering
  \renewcommand{\arraystretch}{1.15}
  \setlength{\tabcolsep}{1.5mm}
  \begin{tabular}{lccccccc}
    \toprule
    Dataset & Audio & mAP $\uparrow$ & FSLA $\uparrow$ & FSLAn $\uparrow$ & FSLAs $\uparrow$ & FSLAm $\uparrow$ \\
    \midrule
    \multirow{4}{*}{FAIR-Play}
    & avg & 35.14 & 54.90 & 88.24 & 63.29 & 39.01 \\
    & L/L & 33.80 & 54.96 & 84.71 & 63.53 & 39.02 \\
    & R/R & 37.27 & 55.21 & 87.06 & 64.46 & 38.02 \\
    & L/R & \textbf{37.63} & \textbf{62.90} & \textbf{89.41} & \textbf{68.63} & \textbf{51.85} \\
    \midrule
    \multirow{4}{*}{BiAVIS-Bench}
    & avg & 22.95 & 43.07 & 32.69 & 47.74 & 13.92 \\
    & L/L & 21.59 & 42.43 & 33.65 & 47.81 & 17.84 \\
    & R/R & 22.91 & 42.21 & 33.65 & 45.96 & 18.83 \\
    & L/R & \textbf{26.65} & \textbf{45.90} & \textbf{60.58} & \textbf{48.14} & \textbf{24.78} \\
    \bottomrule
  \end{tabular}
\end{table*}

\subsection{Limitations}
\label{sec.limitation}
Despite the promising results, the proposed BiAVISM still has three main limitations.
First, compared with monaural AVIS methods, BiAVISM has a stronger dependence on the relative geometry between the camera and the binaural recording device, since the audio-only sound source localization network learns to project binaural cues onto the image plane under a relatively fixed camera--microphone pose.
Consequently, substantial changes in this relative pose may degrade localization accuracy and require fine-tuning, as confirmed by the cross-scenario and camera--microphone geometry evaluations in the supplementary material.
Second, the current formulation focuses on frame-level sounding-instance segmentation to isolate the contribution of binaural spatial cues, without explicitly modeling temporal instance association.
Extending BiAVIS to temporal instance segmentation remains an important next step and requires additional modeling of cross-frame identity association and temporal consistency. A video-level tracking evaluation of the frame-level predictions is reported in the supplementary material.
Third, while BiAVIS-Bench covers a more diverse range of scenes than FAIR-Play, its scale and category coverage remain limited compared with large-scale image or video segmentation datasets. 
This may restrict the model's generalization to unseen object categories and more complex acoustic environments. 
\section{Conclusion and Future Work}
This section summarizes the main findings of this study and discusses its current limitations, broader implications, and future research directions.
This paper studied binaural audio-visual instance segmentation (BiAVIS), which aims to identify and segment visible sounding instances from synchronized RGB frames and human-like binaural audio.
Unlike monaural AVS pipelines that mainly exploit cross-modal semantic correspondence, BiAVIS emphasizes physically grounded spatial cues carried by interaural differences and direction-dependent acoustic filtering.
To support this setting, we annotated FAIR-Play with instance-level sounding masks and collected BiAVIS-Bench using a binocular camera and a human-like binaural microphone in diverse indoor, outdoor, and semi-open scenes.
We further proposed BiAVISM, which first learns image-plane spatial and semantic priors through an audio-only sound source localization branch, and then injects these priors into a query-based instance segmentation decoder via query-level audio--visual fusion.

The work also has clear weaknesses.
As discussed in Section~\ref{sec.limitation}, BiAVISM currently assumes a fixed camera--microphone geometry, so changes in device layout or relative pose may require fine-tuning.
In addition, although BiAVIS-Bench is more diverse than FAIR-Play, its scale and category coverage remain limited relative to large-scale vision datasets, which can constrain generalization to rare objects and complex acoustic mixtures.
From a deployment perspective, RGB frames and binaural audio may contain identifiable faces and voices; therefore, dataset release and model use must remain consent-aware and privacy-preserving, as outlined in the supplementary material.

Several directions remain open for future work.
One priority is calibration-aware or calibration-free binaural learning, so that models can transfer across different camera--microphone baselines, head-related filtering characteristics, and recording devices without expensive re-annotation.
A second direction is to enlarge the benchmark with more categories, denser multi-source overlap, and stronger reverberation and noise.
Finally, although the present study focuses on frame-level BiAVIS to validate the benefit of binaural spatial cues for resolving instance-level intra-class ambiguity, extending it to temporal instance segmentation remains an important next step.

\bibliographystyle{elsarticle-num}

\bibliography{cas-refs}



\end{document}